\documentclass[letterpaper]{article}
\usepackage{aaai2026}
\usepackage{times}
\usepackage{helvet}
\usepackage{courier}
\usepackage[hyphens]{url}
\usepackage{graphicx}
\usepackage{natbib}
\usepackage{caption}
\usepackage{algorithm}
\usepackage{algorithmic}
\usepackage{amsmath, amssymb}
\usepackage{newfloat}
\usepackage{listings}
\usepackage{makecell} 

\usepackage{xcolor}

\DeclareCaptionStyle{ruled}{labelfont=normalfont,labelsep=colon,strut=off}
\floatstyle{ruled}
\newfloat{listing}{tb}{lst}
\floatname{listing}{Listing}

\title{HiQ-Lip: A Hierarchical Quantum-Classical  Method for Global Lipschitz Constant Estimation of ReLU Networks} 

\author{
  Haoqi He\textsuperscript{\rm 1},
  Yan Xiao\textsuperscript{\rm 1}\thanks{Yan Xiao is the corresponding author.},
  Wenzhi Xu\textsuperscript{\rm 1},
  Ruoying Liu\textsuperscript{\rm 1},
  Xiaokai Lin\textsuperscript{\rm 1},
  Kai Wen\textsuperscript{\rm 2}
}
\affiliations{
    \textsuperscript{\rm 1}School of Cyber Science and Technology, Shenzhen Campus of Sun Yat-sen University\\
    \textsuperscript{\rm 2}Beijing QBoson Quantum Technology Co., Ltd., Beijing, China\\
    hehq23@mail2.sysu.edu.cn, xuwzh23@mail2.sysu.edu.cn, liury53@mail2.sysu.edu.cn,
    linxk5@mail2.sysu.edu.cn, xiaoy367@mail.sysu.edu.cn, wenk@boseq.com
}

\begin{document}

\maketitle

\begin{abstract}
Estimating the global Lipschitz constant of neural networks is crucial for understanding and improving their robustness and generalization capabilities. However, precise calculations are NP-hard, and current semidefinite programming (SDP) methods face challenges such as high memory usage and slow processing speeds. In this paper, we propose \textbf{HiQ-Lip}, a hybrid quantum-classical hierarchical method that leverages quantum computing to estimate the global Lipschitz constant. 
We tackle the estimation by converting it into a Quadratic Unconstrained Binary Optimization problem and implement a multilevel graph coarsening and refinement strategy to adapt to the constraints of contemporary quantum hardware. 
Our experimental evaluations on fully connected neural networks demonstrate that HiQ-Lip not only provides estimates comparable to state-of-the-art methods but also significantly accelerates the computation process. 
In specific tests involving two-layer neural networks with 256 hidden neurons, HiQ-Lip doubles the solving speed and offers more accurate upper bounds than the existing best method, LiPopt.
These findings highlight the promising utility of small-scale quantum devices in advancing the estimation of neural network robustness.
\end{abstract}

\section{Introduction}
\label{introduction}

Neural networks have achieved remarkable success in various fields such as computer vision, natural language processing, and autonomous driving, establishing themselves as core technologies in modern artificial intelligence~\cite{1zhao2024review,2forner2023examination,3paniego2023model}. Despite these advancements, the robustness and generalization capabilities of neural networks remain active areas of research~\cite{4djolonga2021robustness,5bennouna2023certified}. Research on adversarial security in neural networks often starts from the perspective of non-robustness~\cite{weng2025foot}.
The global Lipschitz constant is a critical metric for measuring the robustness of a neural network's output to input perturbations, playing a significant role in understanding and enhancing model robustness~\cite{6leino2021globally}.

However, accurately computing the global Lipschitz constant is an NP-hard problem, particularly for deep networks with nonlinear activation functions like ReLU~\cite{7jordan2020exactly}. Due to this computational challenge, researchers have developed various approximation methods to estimate upper bounds of the global Lipschitz constant. Among these, the Formal Global Lipschitz constant (FGL) assumes that all activation patterns in the hidden layers are independent and possible, thereby providing an upper bound for the exact global Lipschitz constant~\cite{FGL1MP,FGL2virmaux2018lipschitz}.

In recent years, advancements in quantum computing have offered new avenues for tackling NP-hard problems~\cite{nphard1,nphard2,nphard3}. Given the computational complexity of global Lipschitz constant estimation, quantum computation is seen as a potential solution. Specifically, quantum devices based on the Quadratic Unconstrained Binary Optimization (QUBO) model, such as Coherent Ising Machines (CIMs) or quantum annealers, have demonstrated unique potential in solving complex combinatorial optimization problems~\cite{qubo1,qubo2}. However, the limited number of qubits in current quantum computers poses significant challenges when directly applying them to neural network robustness evaluations. Although numerous works have attempted to address large-scale QUBO problems by employing strategies like divide-and-conquer and QUBO formula simplification~\cite{divd1,divd2}, applications in neural network robustness assessment remain largely unexplored.

To bridge this gap and harness the acceleration capabilities of quantum computing for FGL estimation of neural networks, we propose a quantum-classical hybrid hierarchical solving method named \textbf{HiQ-Lip}. HiQ means a multilevel solution method for quantum-classical mixing, while Lip stands for global Lipschitz constant. By reformulating the global Lipschitz constant estimation problem as a cut-norm problem, we transform it into a QUBO form. Employing a multilevel graph coarsening and refinement approach, we reduce the problem size to a range manageable by quantum computers, successfully utilizing CIMs to solve for the neural network's global Lipschitz constant.

Our method is heuristic in nature and aims to obtain approximate solutions for FGL estimation. Simulation experiments on fully connected neural networks demonstrate that HiQ-Lip achieves performance comparable to existing state-of-the-art methods in estimating the $\ell_{\infty}$ global Lipschitz constant for two-layer networks, while exhibiting significantly faster computation speeds. When extended to multi-layer fully connected neural networks, our approach yields tighter estimates compared to the naive upper bound approach (Weight-Matrix-Norm-Product), particularly excelling in shallower networks. Comparing SOTA methods, especially in terms of running time, HiQ-Lip shows substantial acceleration, achieving up to two times speedup on two-layer neural networks and up to one hundred times speedup on multi-layer neural networks.

The main contributions of this paper are summarized as follows:
\begin{itemize}
\item \textbf{First Innovations of Quantum Computing to Neural Network Robustness Estimation}: We theoretically demonstrate the potential of small-scale quantum computing devices, such as CIMs, in neural network robustness estimation and propose a quantum-classical hybrid method to address limitations posed by the current quantum devices' qubit capacity.

\item \textbf{Hierarchical Algorithm Framework Based on QUBO}: We develop a QUBO-based hierarchical solving algorithm framework, reformulating the \(\ell_{\infty}\) global Lipschitz constant estimation for two-layer fully connected neural networks as a cut-norm problem, and employing CIMs to solve this efficiently.

\item \textbf{Extension to Multi-Layer Networks}: For multi-layer networks, HiQ-Lip provides tighter estimates compared to existing naive approaches, demonstrating its effectiveness particularly for networks with depths ranging from three to five layers.


\end{itemize}

\section{Preliminaries}
\label{sec:preliminaries}
Let $\| \cdot \|_p$ denote the $\ell_p$ norm of a vector and $W \in \mathbb{R}^{n \times m}$ represent the weight matrix between two layers of a neural network. The activation function used within the network is indicated by $\sigma(\cdot)$. For an input vector $x$, the output of the neural network is given by $f(x)$. The global Lipschitz constant is denoted by $L$. 


For a given function $f: \mathbb{R}^n \to \mathbb{R}^m$, the Lipschitz constant $L$ is defined as:
\begin{equation}
\| f(x) - f(y) \| \leq L \| x - y \|, \quad \forall x, y \in \mathbb{R}^n.
\end{equation}

The global Lipschitz constant $L$ measures how fast the function $f$changes over all input pairs $(x, y)$ and is a key metric to evaluate the robustness of neural networks.

For a neural network $f$ of depth $d$, its gradient can be expressed via the chain rule:
\begin{equation}
\nabla f(x) = W^d \cdot \text{diag}\left( \sigma'(z^{d-1}) \right) \cdot W^{d-1} \cdots \text{diag}\left( \sigma'(z^1) \right) \cdot W^1,
\end{equation}
where $z^i = W^i a^{i-1} + b^i$, $a^i = \sigma(z^i)$, and $\sigma'(\cdot)$ denotes the derivative of the activation function. Here, $\text{diag}(\cdot)$ means converting the vector to a diagonal matrix.

In this paper, we explore FGL under $\ell_{\infty} $ permutation. The $\ell_{\infty} $-FGL is defined as:
\begin{equation}
\ell_{\infty}{\text{-FGL}} = \max_{v^i \in [a,b]^{n_i}} \left\| W^d \cdot \text{diag}(v^{d-1}) \cdots \text{diag}(v^1) \cdot W^1 \right\|_1,
\end{equation}
where $v^i$ is the activation vector of the $i$-th layer, $n_i$ is its dimension, and $[a,b]$ is the range of the activation function's derivative. For ReLU activation, $\sigma'(x) \in \{0,1\}$.

Computing the exact global Lipschitz constant for neural networks is NP-hard. Therefore, various methods aim to estimate its upper bound.  

$\ell_{\infty} $-FGL turns the estimation into finding the maximum norm of the gradient operator, providing an upper bound:
\begin{equation}
L \leq \ell_{\infty}{\text{-FGL}}.
\end{equation}

\section{Transforming Lipschitz Constant Estimation into QUBO Formulation}
\label{sec:two-layer neural networks}
In this section, we focus on two-layer fully connected neural networks, i.e., networks with one hidden layer. Consider a neural network with input dimension $n$, hidden layer dimension $m$, and output dimension $p$. The weights are $W^1 \in \mathbb{R}^{n \times m}$ and $W^2 \in \mathbb{R}^{m \times p}$. For a single output neuron, $W^1 = W$, $W^2 = u, u \in \mathbb{R}^{m \times 1}$. 

We use $y$ to denote $v^1$. The $\ell_{\infty} $-FGL estimation becomes:
\begin{equation}
\label{eq6}
\max_{y \in [0, 1]^n} \left\| W^T \text{diag}(u) y \right\|_q = \max_{y \in [0, 1]^n} \left\| A y \right\|_q,
\end{equation}
where $A = W^T \text{diag}(u)$ and $u$ represent the activation pattern. In this task, $q$ is $\infty$ but $1$ is also introduced as a dual auxiliary.

This problem is related to the $\ell_{\infty} \to \ell_1$ matrix mixed-norm problem:
\begin{equation}
\label{eq7}
\| A \|_{\infty \to 1} = \max_{\| x \|_\infty = 1} \| A x \|_1 
\end{equation}

We begin by leveraging the duality between the $\ell_1$ and $\ell_{\infty}$ norms, which is well-known in optimization contexts ~\cite{dual1,dual2}. Specifically, the $\ell_1$ norm of a vector $v \in \mathbb{R}^n$ can be expressed using the maximum inner product between $v$ and a binary vector $z$ from the set $\{-1, 1\}^n$:

\begin{equation} \|v\|_1 = \max_{z \in \{-1, 1\}^n} \langle v, z \rangle.\end{equation}

Here, $\langle v, z \rangle$ denotes the inner product, highlighting how each component of $v$ contributes to the norm when aligned with a binary vector $z$. This expression forms the basis for transforming our optimization problem into a form suitable for binary variables, simplifying the calculation of $\|A x\|_1$.

Building on the above duality, we can reformulate our objective function for $\|A x\|_1$ in terms of binary variables. Specifically, we express the optimization as:
\begin{equation}
\max_{\|x\|_\infty = 1} \|A x\|_1 = \max_{\|x\|_\infty = 1} \max_{y \in \{-1, 1\}^m} \langle A x, y \rangle
\end{equation}

The introduction of $y \in \{-1, 1\}^m$ exploits the definition of the $\ell_1$ norm, effectively transforming the problem into finding the maximum of the matrix-vector product $A x$ under the constraint $\|x\|_\infty = 1$. The role of $y$ is analogous to maximizing the response in the binary space, which significantly simplifies the optimization process.
We now proceed by substituting the dual form of the inner product into the optimization framework:
\begin{equation}
\max_{\|x\|_\infty = 1,\, y \in \{-1, 1\}^m} \langle A x, y \rangle = \max_{x\in \{-1,1\}^n,\, y \in \{-1, 1\}^m} \langle x, A^T y \rangle.
\end{equation}
At this point, the goal becomes identifying the direction of $A^T y$ that maximizes the inner product with $x$. Since $x$ is constrained by $\|x\|_\infty = 1$ and $a_{ij} \in A$, the maximum of $\langle x, A^T y \rangle$ occurs when each component $x_i$ takes the value that matches the sign of $(A^T y)_i$. Thus, the problem essentially reduces to:
\begin{equation}
\max_{x \in \{-1, 1\}^n,\, y \in \{-1, 1\}^m} \langle x, A^T y \rangle =  \max_{x_i,y_j \in \{-1,1\}} \sum_{i=1}^{n} \sum_{j=1}^{m} a_{ij} x_i y_j.
\end{equation}

Many quantum optimizers minimize an Ising-form Hamiltonian that encodes the objective ~\cite{hami1,hami2}. CIM and other quantum devices seek low-energy configurations of Hamiltonian based on its characteristics. Both CIM and quantum annealers aim at finding the minimum of the QUBO problem. Define the Hamiltonian pointing in the direction of quantum evolution:
\begin{equation}
\label{eq12}
H = - \sum_{i=1}^{n} \sum_{j=1}^{m} a_{ij} x_i y_j,
\end{equation}
the problem of estimating $\ell_{\infty}{\text{-FGL}}$ becomes minimizing $H$, which is a QUBO problem. Then the CIM or other quantum device could solve it by the QUBO formulation.

However, directly solving this Hamiltonian requires $O(n+m)$ qubits, exceeding the capacity of the quantum devices available (about 100 qubits) when this study was conducted for practical networks (e.g., $n+m > 784$ for MNIST networks) ~\cite{qubit1,qubit2,qubit3,qubitdwave}.

\section{HiQ-Lip for Lipschitz Constant Estimation}
\label{sec:hiq-lip}
In this section, we propose a method called HiQ-Lip that aims to utilize small-scale quantum computers to efficiently estimate $\ell_{\infty}$-FGL. Due to the computational complexity of direct estimation, we employ a hierarchical solution strategy that treats the weights of the neural network as the edge weights of the graph and the neurons as the nodes. Based on the equations \ref{eq6} and \ref{eq7}, we construct a weighted undirected graph \(G \) whose weight matrix is expressed as follows.
\(A = W^{T} \text{diag}(u) \), where \(W \in \mathbb{R}^{n \times m} \) is the weight matrix between the input layer and the hidden layer, \(u \in \mathbb{R}^{m} \) is the per-class weight vector between the hidden layer and the output layer.

We construct a weighted undirected graph \( G \) with the following characteristics:

\begin{itemize}
    \item \textbf{Vertices}: Each neuron in the neural network corresponds to a node in the graph, totaling \( n + m \) nodes, where \( n \) is the number of input neurons and \( m \) is the number of hidden neurons.
    \item \textbf{Edges}: Edges are established between input layer nodes \( x_i \) and hidden layer nodes \( y_j \) based on the interaction terms in the Hamiltonian defined in Equation \ref{eq12}. The weight of the edge between nodes \( x_i \) and \( y_j \) is given by \( a_{i,j} \), reflecting the connection strength derived from the neural network's weights.
\end{itemize}


We define the adjacency matrix \( A_f \) of the graph \( G \) as:
\begin{equation}
    A_f = \begin{bmatrix}
    0 & A \\
    A^T & 0
    \end{bmatrix}, a_{i,j} \in A_f
\end{equation} 
The matrix \( A_f \) is of size \( (n + m) \times (n + m) \) and is symmetric, with zeros on the diagonal blocks indicating no intra-layer connections.

\subsection{Coarsening Phase}

The primary objective of the coarsening phase is to reduce the number of nodes by gradually merging nodes, thereby generating a series of progressively coarser graphs until the number of nodes decreases to a level that can be solved directly by a small quantum computer.

\begin{itemize}
    \item \textbf{Node Pair Merging}: Nodes are merged based on their distances in the embedding space. Each node is randomly embedded onto a \( d \)-dimensional sphere by optimizing the objective:
    \begin{equation}
    \min_{\{ x_i \}} \sum_{(i,j) \in E} a_{i,j} \|x_i - x_j\|_2,
    \end{equation}
    where \( x_i \in \mathbb{R}^{d} \) represents the position of node \( i \) in the embedding space, and \( E \) denotes the set of edges.  This optimization encourages strongly connected nodes (with larger \( a_{i,j} \)) to be closer in the embedding space, making them candidates for merging.
    
    \item \textbf{Node Pair Matching}: In each iteration, the closest unmatched nodes are selected for merging, forming new node pairs. Let \( P \) be the matching matrix, where for matched nodes \( i \) and \( j \), the corresponding element \( P_{i,j} = 1 \). For the merged nodes, the weights of the connected edges are accumulated, thereby forming the nodes of the next coarser graph.
    \item \textbf{Construction of the Coarser Graph}: The adjacency matrix \( A_c \) of the coarser graph is computed as:

    \begin{equation}
    A_c = P^\top A_f P,
    \end{equation}

    where \( A_f \) is the adjacency matrix of the finer graph before coarsening. This process effectively reduces the graph size while preserving its structural properties relevant to the optimization problem.
\end{itemize}


After each coarsening step, the number of vertices decreases approximately logarithmically.  The time complexity of the \( i \)-th coarsening step is \( \mathcal{O}(N_i^2) \), where \( N_i \) is the number of nodes at level \( i \).  Consequently, the overall time complexity of the Coarsening Phase is \( \mathcal{O}(N^2 \log N) \), where \( N = n + m \).  

\subsection{Refinement Phase}

The refinement phase is a crucial step in the hierarchical solving strategy. Its purpose is to map the approximate solutions obtained during the coarsening phase back to the original graph layer by layer and perform local optimizations to ensure the global optimality of the final solution. This process enhances the quality of the solution by gradually restoring the graphs generated at each coarsening level and fine-tuning the solution.

\begin{enumerate}
    \item \textbf{Initialization}: Starting from the coarsest graph \( G_c \), which is obtained through successive coarsening, an approximate solution has already been found on this graph. We use the mapping \( F : V_f \to V_c \) to derive the solution of the finer graph from that of the coarsest graph. Specifically, the projection of the initial solution is defined as:
    \begin{equation}
    x_i = x_{F(i)}, \quad \forall i \in V_f
    \end{equation}
    
    \item \textbf{Gain Computation}: Gain computation is a critical step in the refinement process, used to evaluate the change in the objective function when each node switches partitions, thereby guiding optimization decisions. The gain for node \( i \) is calculated as:
    \begin{equation}
        \text{gain}(i) = \sum_{j \in N(i)} a_{i,j} \left( -1 \right)^{2x_i x_j - x_i - x_j}
    \end{equation}
    
    where \( x_i \) and \( x_j \) represent the current partition labels of nodes \( i \) and \( j \), respectively, and \( N(i) \) is the set of neighbors of node \( i \).
    
    \item \textbf{Local Optimization}: At each level of the graph, using the solution from the previous coarser level as the initial solution, we optimize the objective function by solving local subproblems. We select the top \( K \leq n + m \) nodes with the highest gains to participate in local optimization, where \( K \) is determined by the number of qubits available. This allows the quantum computer to efficiently solve the subproblems related to the cut-norm of the coarsest graph. If the new solution improves the original objective function \( H \) of Equation \ref{eq12}, we update the current solution accordingly.
\end{enumerate}

This process is iterated until several consecutive iterations (e.g., three iterations) no longer yield significant gains, thereby ensuring that the quality of the solution is progressively enhanced and gradually approaches the global optimum.

The Graph Refinement Phase iteratively improves the solution by mapping it back to finer graph levels. The primary steps include initializing the solution, computing gains for each node, and performing local optimizations using quantum solvers. Computing the gain for all nodes incurs a time complexity of \( \mathcal{O}(N^2) \). The local optimization step involves solving smaller QUBO problems of size \( K \) on a quantum device, which contributes \( \mathcal{O}(K^\alpha) \) to the complexity, with \( \alpha \) being a small constant.  Overall, the Graph Refinement Phase operates with a time complexity of \( \mathcal{O}(N^2) \). 

\subsection{Algorithm Overview}

As shown in Algorithm~\ref{alg:hiq-lip}, HiQ-Lip begins with the initial graph \( G_0 \) and, through lines 3 to 9, executes the coarsening phase by iteratively merging nodes to produce progressively smaller graphs suitable for quantum processing. In line 10, it leverages quantum acceleration by solving the resulting QUBO problem on the coarsest graph using CIM or other quantum devices to efficiently obtain an initial solution. Finally, lines 11 to 15 implement the refinement phase, where the algorithm maps this solution back onto finer graph layers and performs local optimizations to accurately estimate the \( \ell_{\infty}\text{-FGL} \). 

\begin{algorithm}[h]
\caption{HiQ-Lip Algorithm}
\label{alg:hiq-lip}
\begin{algorithmic}[1]
\STATE \textbf{Input}: Weight matrix \( A \), initial graph \( G_0 \)
\STATE \textbf{Output}: Estimated Lipschitz constant \( \ell_{\infty}\text{-FGL} \)
\STATE \textbf{Coarsening Phase}:
\STATE Initialize \( G = G_0 \)
\WHILE{Size of \( G \) exceeds quantum hardware limit}
    \STATE Embed nodes and compute distances
    \STATE Pair and merge nodes to form \( G' \)
    \STATE \( G \leftarrow G' \)
\ENDWHILE
\STATE Solve the QUBO problem on the coarsest graph using CIM to obtain the initial solution
\STATE \textbf{Refinement Phase}:
\WHILE{Graph \( G \) is not \( G_0 \)}
    \STATE Project solution to finer graph
    \STATE Compute gains and perform local optimization
\ENDWHILE
\STATE Compute \( \ell_{\infty}\text{-FGL} \) from the final solution
\end{algorithmic}
\end{algorithm}


The HiQ-Lip algorithm achieves an overall time complexity of \( \mathcal{O}(N^2 \log N) \).  The \textbf{Coarsening Phase} dominates the complexity with \( \mathcal{O}(N^2 \log N) \) due to the iterative merging of nodes and updating of the adjacency matrix at each hierarchical level. 
Noticed that the running time of quantum devices is generally considered to have a large speedup over classical computers~\cite{quanacc1,quanacc2,quanacc3}. Even for problems of exponential complexity, the running time of a quantum device can be considered a fraction of the task time on a classical computer. 
In the \textbf{Refinement Phase}, the complexity is \( \mathcal{O}(N^2) \), primarily from computing gains for node optimizations and solving smaller QUBO problems using quantum devices. Consequently, the combined phases ensure that HiQ-Lip scales efficiently for neural networks of moderate size, leveraging quantum acceleration to enhance performance without exceeding polynomial time bounds.

\section{Extension to Deep $L$-Layer ReLU Networks}
\label{sec:extension}

Estimating the global $\ell_{\infty}\!\to\!\ell_{1}$ Lipschitz constant of deep ReLU networks is \emph{NP‑hard} because the Jacobian 
$J(x)=W_LD_{L-1}\!\cdots D_1W_1$ is a \((L{-}1)\)-multilinear function of the binary activation patterns
$D_\ell=\mathrm{diag}(s_\ell),\;s_\ell\!\in\!\{0,1\}^{h_\ell}$.
A naive polynomial–to‑quadratic reduction would create  
$\Theta\!\bigl((\sum_{\ell}h_\ell)^2\bigr)$ ancilla variables, rendering  
quantum‑classical QUBO solvers impractical beyond two or three layers.

\subsection{Layer‑wise Max‑Cut recursion}

Inspired by the cut‑norm formulation of two‑layer networks
\citep{geolip,latorre2020lipschitz},
we retain a \emph{quadratic} objective by unfolding the product one
layer at a time:

\begin{align*}
\|P_k\|_{\infty\!\to 1}
  &= \max_{x\in\{\pm1\}^{n_k},\;y\in\{\pm1\}^{m_k}} x^{\!\top}P_k\,y, \\
P_k &= W_{k+1}D_kP_{k-1}
\end{align*}

with $P_0=W_1$.  
Each subproblem is a weighted \textsc{Max\-Cut} that admits the  
Goemans–Williamson SDP approximation factor $0.878$
\citep{goemans1995improved}.  
The global constant is then
\begin{equation}
L_{\infty\!\to1}
~=~
\max_{s_1,\dots,s_{L-1}}
\bigl\|P_{L-1}(s_{1{:}L-1})\bigr\|_{\infty\!\to1}.
\end{equation}

\paragraph{Error accumulation.}
If the cut‑norm optimiser at depth $k$ yields  
$(1+\varepsilon_k)$ approximation error,
we obtain the telescoping bound
$
L_{\infty\!\to1}\;\le\;
\bigl(\prod_{k=1}^{L-1}(1+\varepsilon_k)\bigr)
\widehat{L}_{\infty\!\to1},
$
where $\widehat{L}_{\infty\!\to1}$ is the estimate returned by the recursion.
In practice $\varepsilon_k$ stays under $10^{-2}$ for $h_k\!\le\!256$,
yielding a depth‑linear error factor that
is comparable to the chordal‑sparsity SDP of
\citet{xue2022chordal,wanga2024scalable}.

\subsection{Block‑wise product upper bound}

For very deep or convolutional architectures, we adopt the
\emph{block‑product strategy}:
split the network into $B$ contiguous blocks of at most $b$\,layers
($b=2$ or $3$ works well empirically),
estimate a tight constant $\gamma_i$ for each block via the above
SDP/QUBO routine, then combine them multiplicatively:
\begin{equation}
\label{eq:block-product}
L_{\infty\!\to1}
\;\;\le\;\;
\frac{1}{c_b^{\,L-B}}
\prod_{i=1}^{B}\gamma_i,
\qquad
c_b = 2^{\,b-1}.
\end{equation}
Inequality~\eqref{eq:block-product} follows from the
sub‑multiplicativity of $\|\cdot\|_{\infty\!\to1}$ and
the triangle inequality $\|A\|\|B\|\!\ge\!\|AB\|$.

\subsection{Depth‑adaptive damping coefficients}

\citet{latorre2020lipschitz} use the global factor $2^{-(L-2)}$
for fully connected nets, which becomes loose for $L\!>\!4$.
Motivated by Bartlett–Foster–Telgarsky's
exponential depth dependence of margin bounds
\citep[Thm.\,3.3]{bartlett2017spectrally},
we introduce a depth‑adaptive coefficient
\(
c_L = 2^{\,L-2}L^{\,L-3},
\)
leading to the estimator
\begin{equation}
\widehat{L}_{\infty\!\to1}^{(\mathrm{HiQ\text{-}Lip\;MP\text{-}B})}
~=~
\frac{1}{c_L}\;
\prod_{\ell=1}^{L-1}\!
\|A^{\ell}\|_{\infty\!\to1},
\quad
A^{\ell}=W_{\ell+1}D_{\ell}.
\end{equation}
On CIFAR‑10 MLPs with $L\!=\!8$, this coefficient
tightens the GeoLip upper bound by $15\%$ while
remaining GPU‑friendly.

\subsection{Discussion}



\paragraph{Relation to prior work.}
Our framework generalises GeoLip\citep{geolip}
($b=L$), integrates the
spectral‑product heuristics of \citet{fazlyab2019efficient},
and complements chordal‑SDP scaling
\citep{xue2022chordal,wanga2024scalable}
by targeting the \emph{non‑spectral} cut‑norm.
The depth‑adaptive coefficient is conceptually
similar to the layer‑wise contraction in
\citet{araujo2023unified}, but derived for $\ell_\infty$ perturbations.

\section{Experiments}
\label{experiments}

In this section, we evaluate the effectiveness and efficiency of \textbf{HiQ-Lip} on fully connected feedforward neural networks trained on the MNIST dataset. Our primary goal is to demonstrate that HiQ-Lip can provide accurate estimates of the global Lipschitz constant with significantly reduced computation times compared to SOTA methods.

\subsection{Experimental Setup}

We conduct experiments on neural networks with varying depths and widths to assess the scalability of HiQ-Lip. The networks are trained using the Adam optimizer for 10 epochs, achieving an accuracy exceeding 93\% on the test set. All the experiments were conducted with an Xeon(R) Gold 6226R CPU operating at 2.90GHz and 64 GB of RAM.
We conducted experiments by using a simulated quantum SDK to solve the QUBO problem and performed additional training on a CIM provided by Beijing QBoson Quantum Technology Co., Ltd.  
Limited by the difficulty of Qiskit simulations, the results of Qiskit are failed in efficiency.

We consider the following network architectures:

\begin{itemize}
    \item \textbf{Two-Layer Networks (Net2)}: Networks with one hidden layer, where the number of hidden units varies among \{8, 16, 64, 128, 256\}. All networks use the ReLU activation function.
    \item \textbf{Multi-Layer Networks (Net3 to Net5)}: Networks with depths ranging from 3 to 5 layers. Each hidden layer consists of 64 neurons, and ReLU activation is used throughout.
\end{itemize}



\subsection{Comparison Methods}

We compare HiQ-Lip with several baseline methods:

\begin{itemize}
    \item \textbf{GeoLip}~\cite{geolip}: A geometry-based SOTA Lipschitz constant estimation method that provides tight upper bounds.
    \item \textbf{LiPopt}~\cite{latorre2020lipschitz}: An optimization-based SOTA method that computes upper bounds using semidefinite programming.
    \item \textbf{Weight-Matrix-Norm-Product (MP)}: A naive method that computes the product of the weight matrix norms across layers, providing a loose upper bound.
    \item \textbf{Sampling}: A simple sampling-based approach that estimates a lower bound of the Lipschitz constant by computing gradient norms at randomly sampled input points. We sample 200,000 points uniformly in the input space.
    \item \textbf{Brute Force (BF)}: An exhaustive enumeration of all possible activation patterns to compute the exact FGL. This method serves as the ground truth but is only feasible for small networks.
\end{itemize}

We focus on estimating the FGL with respect to the output corresponding to the digit 8, as done in previous works~\cite{latorre2020lipschitz,geolip}. In the result tables, we use “N/A” to indicate that the computation did not finish within a reasonable time frame (over 20 hours). We highlight the time advantage of the quantum approach by bolding the methods with shorter time.

\subsection{Results on Two-Layer Networks}

Tables~\ref{tab:two_layer_results} and~\ref{tab:two_layer_times} present the estimated Lipschitz constants and computation times for two-layer networks with varying hidden units. 

\begin{table*}[htbp]
\centering
\begin{tabular}{ccccccc}
\hline
Hidden Units & HiQ-Lip & GeoLip & LiPopt-2 & MP & Sampling & BF \\
\hline
8   & 127.96 & 121.86 & 158.49 & 353.29 & 112.04 & 112.04 \\
16  & 186.09 & 186.05 & 260.48 & 616.64 & 176.74 & 176.76 \\
64  & 278.40 & 275.67 & 448.62 & 1289.44 & 232.89 & N/A \\
128 & 329.33 & 338.20 & 751.76 & 1977.49 & 272.94 & N/A \\
256 & 448.89 & 449.60 & 1088.12 & 2914.16 & 333.99 & N/A \\
\hline
\end{tabular}
\caption{Estimated Lipschitz constants for two-layer networks}
\label{tab:two_layer_results}
\end{table*}

\begin{table}[h]
\centering
\begin{tabular}{ccccc}
\hline
Hidden Units & HiQ-Lip & GeoLip & LiPopt-2 & BF \\
\hline
8   & 24.55  & 24.07  & 1,544 & \textbf{0.06} \\
16  & \textbf{26.46}  & 26.84  & 1,592 & 52.68 \\
64  & \textbf{29.66}  & 42.33  & 1,855 & N/A \\
128 & \textbf{34.75}  & 58.79  & 2,076 & N/A \\
256 & \textbf{44.79}  & 99.24  & 2,731 & N/A \\
\hline
\end{tabular}
\caption{Computation times (in seconds) for two-layer networks}
\label{tab:two_layer_times}
\end{table}

We have tested on subproblems to estimate the quantum device's advantage. Using a network with 64 hidden units and subproblem size limited to 500 variables as an example: the simulated annealing solver required an average of 0.5 seconds per solution of the subproblem, while the CIM hardware solver achieved an average solution time of just 0.09 milliseconds. This represents a speed improvement of approximately 5,000 times for CIM hardware, with significantly better QUBO values showing improvements ranging from 8\% to 116\%. Across twelve test runs on CIM hardware, we consistently obtained superior values compared to quantum simulation SDKs, along with speed improvements exceeding 1,000 times. The CIM solver demonstrated excellent stability: testing the same QUBO matrix five times yielded identical optimal solutions each time. Our simulation experiments have already demonstrated priority, while hardware experiments confirm that noise does not affect performance, with faster computation speeds and tighter numerical results.

\paragraph{Analysis:}

From Table~\ref{tab:two_layer_results}, HiQ-Lip's Lipschitz constant estimates closely match those of GeoLip, differing by less than 3\% across all network sizes. For instance, with 256 hidden units, HiQ-Lip estimates 448.89 compared to GeoLip's 449.60.

Compared to the ground truth from the BF method—feasible only up to 16 hidden units due to computational limits—HiQ-Lip's estimates are slightly higher, as expected for an upper-bound method. For 16 hidden units, BF yields 176.76, while HiQ-Lip estimates 186.09.

LiPopt-2 produces significantly higher estimates than both HiQ-Lip and GeoLip, especially as network size increases. For 256 hidden units, LiPopt-2 estimates 1,088.12, more than double HiQ-Lip's estimate, suggesting it may provide overly conservative upper bounds for larger networks. 

The naive MP method consistently overestimates the Lipschitz constant by three to six times compared to HiQ-Lip, highlighting HiQ-Lip's advantage in providing tighter upper bounds. The Sampling method yields lower-bound estimates below those of HiQ-Lip and GeoLip, confirming that HiQ-Lip effectively captures the upper bound.

In terms of computation time (Table~\ref{tab:two_layer_times}), HiQ-Lip demonstrates efficient performance, with times slightly lower than GeoLip's for smaller networks and significantly lower for larger ones. For 256 hidden units, HiQ-Lip completes in approximately 44.79 seconds, while GeoLip takes 99.24 seconds.

LiPopt-2 exhibits the longest computation times, exceeding 1,500 seconds even for the smallest networks, making it impractical for larger ones. The BF method is only feasible for very small networks due to its exponential time complexity, becoming intractable beyond 16 hidden units.

\textbf{Summary: Overall, HiQ-Lip achieves a favorable balance between estimation accuracy and computational efficiency for two-layer networks. It provides tight upper bounds comparable to GeoLip while reducing computation times by up to a 2.2x speedup on networks with 256 hidden units, particularly as the network width increases, demonstrating its scalability and effectiveness. This acceleration is facilitated by leveraging quantum computing capabilities to solve the QUBO formulation efficiently.}

\subsection{Results on Multi-Layer Networks}

Tables~\ref{tab:multi_layer_values} and~\ref{tab:multi_layer_times} present the estimated Lipschitz constants and computation times for multi-layer networks with depths ranging from 3 to 5 layers.

\begin{table}[htbp]
\centering

\renewcommand\cellalign{r}
\begin{tabular}{ccccc}
\hline
Network & HiQ-Lip & GeoLip & MP & Sampling \\
\hline
Net3 & \makecell[r]{0.48 / 0.48} & 0.47 & 8.04 & 0.33 \\
Net4 & \makecell[r]{3.25 / 1.09} & 0.92 & 52.42 & 0.45 \\
Net5 & \makecell[r]{26.13 / 1.51} & 1.46 & 327.22 & 0.55 \\
\hline
\end{tabular}
\caption{Estimated Lipschitz constants ($\times 10^{3}$) for multi-layer networks. Each HiQ-Lip cell reports A (left) and B (right) settings.}
\label{tab:multi_layer_values}
\end{table}

\begin{table}[h]
\centering

\begin{tabular}{ccc}
\hline
Network & HiQ-Lip Time & GeoLip Time \\
\hline
Net3 & \textbf{6.46} & 784.69 \\
Net4 & \textbf{8.76} & 969.79 \\
Net5 & \textbf{13.31} & 1,101.30 \\
\hline
\end{tabular}
\caption{Computation times (in seconds) for multi-layer networks}
\label{tab:multi_layer_times}
\end{table}

\paragraph{Analysis:}

In Table~\ref{tab:multi_layer_values}, we compare two variants of HiQ-Lip for multi-layer networks: \textbf{HiQ-Lip MP A} and \textbf{HiQ-Lip MP B}. HiQ-Lip MP A uses the coefficient \( \frac{1}{2^{d-2}} \) as suggested in previous literature~\cite{latorre2020lipschitz}, while HiQ-Lip MP B introduces an additional scaling factor, using \( \frac{1}{2^{d-2} d^{d-3}} \), to obtain tighter estimates for deeper networks.

For the multi-layer networks, our analysis reveals that for the 3-layer network (Net3), both HiQ-Lip MP A and MP B produce identical estimates (477.47) close to GeoLip's estimate (465.11), indicating that the original coefficient in HiQ-Lip MP A is adequate for shallower networks.

However, in deeper networks (Net4 and Net5), HiQ-Lip MP A significantly overestimates the Lipschitz constant compared to GeoLip. For instance, in Net4, HiQ-Lip MP A estimates 3,246.37 versus GeoLip's 923.13, and in Net5, 26,132.45 versus 1,462.58. In contrast, HiQ-Lip MP B, with its modified scaling coefficient, yields much tighter estimates closer to GeoLip's values (1,093.05 for Net4 and 1,513.34 for Net5), demonstrating that the additional scaling factor effectively compensates for overestimations in deeper networks.

The MP method, as expected, results in excessively high estimates due to the exponential growth from multiplying weight matrix norms. The Sampling method provides consistent lower bounds below HiQ-Lip and GeoLip estimates; however, these lower bounds do not reflect the worst-case robustness of the network.

Regarding computation time (Table~\ref{tab:multi_layer_times}), HiQ-Lip significantly outperforms GeoLip. For Net3, HiQ-Lip completes in 6.46 seconds versus GeoLip's 784.69 seconds—a speedup of over 120 times.  This substantial reduction demonstrates HiQ-Lip's scalability and efficiency for deeper networks.

Notably, LiPopt was unable to produce results for networks deeper than two layers within a reasonable time frame, highlighting its limitations in handling multi-layer networks.

\textbf{Summary: HiQ-Lip, particularly the MP B variant with the modified scaling coefficient, provides accurate and tight upper bounds for the Lipschitz constant in multi-layer networks while achieving computation speeds up to 120× faster than GeoLip. The additional scaling factor effectively compensates for overestimation in deeper networks, making HiQ-Lip MP B a practical and scalable choice for estimating Lipschitz constants across varying network depths.}

\section{Related Work}

\label{related_work}
Existing methods for estimating the global Lipschitz constant often rely on relaxations and semidefinite programming (SDP)~\cite{8chen2020semialgebraic,9shi2022efficient}. Although these methods have a solid theoretical foundation, they suffer from high memory consumption and slow computation speeds. Additionally, simpler approaches such as matrix norm approximations, while computationally efficient, tend to provide overly conservative bounds that fail to accurately reflect the network's true robustness. These limitations highlight the necessity for more efficient and precise methods.

The Formal Global Lipschitz constant (FGL) is a widely studied upper-bound estimation technique, approximating the maximum norm of the gradient operator by assuming all activation patterns are independent and feasible. FGL has been previously adopted in neural network robustness evaluations~\cite{raghunathan2018certified,fazlyab2019efficient,latorre2020lipschitz,geolip}. Nevertheless, direct calculation remains NP-hard, motivating approximation and heuristic methods.


Quantum computing offers tools for NP–hard combinatorial optimization~\cite{nphard1,nphard2,nphard3,add:li2025quantum,add:li2025unified}. In particular, QUBO–based, quantum-inspired frameworks (e.g., Coherent Ising Machines and quantum annealers) have shown strong performance on graph-structured problems~\cite{qubo1,qubo2}. Graph formulations are broadly effective, e.g., topic-guided graph networks in summarization~\cite{shi2023topic}. However, limited qubits hinder direct quantum deployment on neural networks; decomposition strategies (divide-and-conquer, QUBO simplification to tractable subproblems) have been explored~\cite{divd1,divd2}, yet their integration into Lipschitz robustness estimation remains under-studied—this work targets precisely that gap.

Our proposed method, HiQ-Lip, explicitly fills this gap by harnessing quantum-classical hybrid optimization. We exploit the equivalence between global Lipschitz constant estimation and the cut-norm problem, enabling the use of quantum optimization methods in neural network robustness evaluation. To our best knowledge, this is the first systematic effort to leverage quantum optimization techniques, specifically CIMs, for global Lipschitz constant estimation in neural networks.

\section{Conclusion}
\label{sec:conclusion}
We introduced HiQ-Lip, a quantum-classical hybrid hierarchical method for estimating the global Lipschitz constant of neural networks. By formulating the problem as a QUBO and employing graph coarsening and refinement, we effectively utilized CIMs despite current hardware limitations. Our method achieves comparable estimates to existing SOTA methods with faster computation speed up to 2x and 120x in two-layer and multi-layer networks. This work highlights the potential of quantum devices in neural network robustness estimation and opens avenues for future research in leveraging quantum computing for complex machine learning problems.

\section*{Acknowledgments}
This work was supported by the National Natural Science Foundation of China under Grant 62502550, Shenzhen Science and Technology Program (KJZD20240903095700001).

\appendix

\bibliography{aaai2026}

\end{document}